\documentclass{article}

\usepackage{microtype}
\usepackage{graphicx}
\usepackage{subfigure}
\usepackage{booktabs} 

\usepackage{multirow}

\usepackage{amsmath,amsfonts,bm}

\def\eqref#1{equation~\ref{#1}}

\def\1{\bm{1}}

\DeclareMathAlphabet{\mathsfit}{\encodingdefault}{\sfdefault}{m}{sl}
\SetMathAlphabet{\mathsfit}{bold}{\encodingdefault}{\sfdefault}{bx}{n}

\usepackage{wrapfig}
\usepackage{hyperref}
\usepackage{url}

\usepackage{booktabs}       
\usepackage{amsfonts}       
\usepackage{nicefrac}       
\usepackage{algorithm}
\usepackage{algorithmic}
\usepackage{times}
\usepackage{graphicx}
\usepackage{epsfig}
\usepackage{amsmath}
\usepackage{amssymb}
\usepackage{caption}
\usepackage{multicol}
\graphicspath{{fig/}}

\newtheorem{theorem}{Theorem}
\newtheorem{definition}{Definition}

\usepackage{hyperref}

\usepackage[accepted]{icml2021}

\icmltitlerunning{A Tale of Two Efficient and Informative Negative Sampling Distributions}

\begin{document}

\twocolumn[
\icmltitle{A Tale of Two Efficient and Informative Negative Sampling Distributions}




\begin{icmlauthorlist}
\icmlauthor{Shabnam Daghaghi}{to}
\icmlauthor{Tharun Medini}{to}
\icmlauthor{Nicholas Meisburger}{goo}
\icmlauthor{Beidi Chen}{ed}
\icmlauthor{Mengnan Zhao}{to}
\icmlauthor{Anshumali Shrivastava}{goo,to}
\end{icmlauthorlist}

\icmlaffiliation{to}{Department of Electrical and Computer Engineering, Rice University}
\icmlaffiliation{ed}{Department of Computer Science, Stanford University}
\icmlaffiliation{goo}{Department of Computer Science, Rice University}

\icmlcorrespondingauthor{Shabnam Daghaghi}{shabnam.daghaghi@rice.edu}

\icmlkeywords{Machine Learning, ICML}

\vskip 0.3in
]



\printAffiliationsAndNotice{} 

\begin{abstract}
Softmax classifiers with a very large number of classes naturally occur in many applications such as natural language processing and information retrieval. The calculation of full softmax is costly from the computational and energy perspective. There have been various sampling approaches to overcome this challenge, popularly known as negative sampling (NS). Ideally, NS should sample negative classes from a distribution that is dependent on the input data, the current parameters, and the correct positive class. Unfortunately, due to the dynamically updated parameters and data samples, there is no sampling scheme that is provably adaptive and samples the negative classes efficiently. Therefore, alternative heuristics like random sampling, 
static frequency-based sampling, or learning-based biased sampling, 
which primarily trade either the sampling cost or the adaptivity of 
samples per iteration are adopted. In this paper, we show two classes of distributions where the sampling scheme is truly adaptive and provably generates negative samples in near-constant time. Our implementation in C++ on CPU is significantly superior, both in terms of wall-clock time and accuracy, compared to the most optimized TensorFlow implementations of other popular negative sampling approaches on powerful NVIDIA V100 GPU.
\end{abstract}

\vspace{-5mm}
\section{Introduction}\label{sec:intro}
\vspace{-1mm}
Neural Networks (NN) have successfully pushed the boundaries of many application tasks, such as image or text classification \citep{wang2017residual,yao2019graph}, speech recognition \citep{dong2018speech} and recommendation systems~\citep{zhang2015character,medini2019extreme}. Many hard AI problems are currently modeled as massive multiclass or multilabel problems leading to a drastic improvement over prior work. For example, popular NLP models predict the best word, given the full context observed so far. Such models are becoming state-of-the-art. Recommendation systems and related Information Retrieval (IR) problems are classical examples of machine learning with outrageously large outputs~\citep{medini2019extreme,jain2019slice}. In IR, given the user query, the task is to predict few relevant documents (or products) from among hundreds of millions of possible documents, a typical machine learning problem with massive output space. 

Owing to the significance of the problem, \emph{machine learning with large output space} or alternatively also known as \emph{extreme classification} is a field in itself~\citep{bengio2019extreme}. A large number of classes naturally brings a new set of computational and memory challenges. 


Fortunately, with access to powerful Graphics Processing Unit (GPU)~\citep{owens2008gpu}, the training processes of large models have been accelerated heavily. That is because GPUs have a unique advantage for matrix multiplication, which usually requires a cubic time algebraic operation ($\mathcal{O}(N^3)$) and is the major and costly building block of NN computations. However, the number of concurrent operations required in large matrix multiplications for classification with an extensive number of classes has reached a limit for further speedups even using GPUs.

\vspace{-2mm}
\subsection{Negative Sampling}
\vspace{-2mm}
The typical approach to address the challenge mentioned above is known as negative sampling~\citep{pennington2014glove,jean2014using,rawat2019sampled, mikolov2013distributed}. In Negative Sampling, we select a small subset of classes for each input and compute the softmax and cross-entropy function. This subset usually includes the positive (true) and a small set of negative (false) classes. Negative sampling reduced the computations in the most cumbersome last layer, thereby making the gradient update procedure efficient.

However, approximating full softmax with small sub-sample results in poor convergence if the negative samples are not chosen appropriately. For instance, let us take the example of a recommendation system (predicting products relevant to a query) with a large number of products. If the input query is `Nike Running Shoes,' the true loss concentrates on the specific small number of confusing ('hard') negative classes like `Adidas Running Shoes'. Since the number of classes is huge, random sampling is unlikely to identify this hard negative class. Other heuristics like frequent class sampling as negative samples are also unlikely to find these hard negatives most of the time. Frequent class sampling will probably choose `iphone' as a potential solid negative sample due to its popularity. Clearly, without discriminating between closely related negative samples, the classifier cannot achieve good accuracy. Our experiments on recommendations datasets clearly indicate this sub-optimality of current negative sampling heuristics. 

If there exists a way to sample the subset of confusing classes from the skewed distribution, the training progress would be largely accelerated. However, as evident from the example, such ground-truth distribution depends on the input sample and current model parameters. Moreover, this distribution varies significantly as training progresses. Consider the same query 'Nike Running Shoes'. Initially, when the network has not learned anything and has random weights, all classes are equally confusing. Thus, uniform sampling is optimal initially as the network has just started to learn. As the training progresses, the network's belief starts getting more concentrated on a few classes; at this time, a negative sample of say 'baby toys' for query `Nike Running Shoes' is not a very informative negative sample because the network has already learned to tell them apart. The sampling distribution keeps changing, often drastically, as the training progresses.

If we have $N$ classes, given the labeled training instance $(x,y)$, the relevance of any other class $z \ne y$ is a non-trivial function of the input and the parameters. To the best of our knowledge, there does not exist any statistical sampling scheme for adaptive Negative Sampling, where the cost of maintaining and updating the distribution, per iteration, is asymptotically $\mathcal{O}(1)$ (independent (or logarithmic) of the number of classes). The input feature $x$, current true class $y$, and the parameters all change every iteration, causing the sampling weights to change. As a result,  it appears that any non-trivial adaptive sampling will require at least $O(N)$ work even to compute these sampling weights (or score). It is widely assumed that there is no such sampling scheme, and hence several heuristic alternatives are proposed.

\textbf{Negative Sampling Heuristic with a Static Distribution:} The first set of alternatives use a static distribution ~\citep{bengio2008adaptive}. The most popular ones, implemented in TensorFlow, assume a static distribution such as the distribution based on the frequency of classes. Uniform sampling is another popular choice.   
\\

\textbf{Fundamental Problem with Learning Based Negative Sampling Heuristic:} Learning-based alternatives are becoming a popular alternative~\citep{bamler2020extreme, pmlr-v9-gutmann10a}. Here, a machine learning generator predicts (or generates) the negative samples. However, it is a chicken-and-the-egg problem. 
The generator is solving the same hard problem, prediction over a large number of classes, as a sub-routine. Note, the size of the outputs $N$, still remains the same even for the learning-based generator. Furthermore, since the sampling distribution for the same data point shifts drastically throughout training because of parameter updates, it is not clear if a learned generator, which ignores the network's parameters values, can produce relevant samples at every iteration of the training. 

Negative sampling alternatives try to balance the sampling cost with quality. So far, negative sampling methods, other than the ones based on static sampling, have failed to demonstrate any training time improvements over the optimized full softmax implementation over GPUs. Static sampling strategies are known to be fast but lead to poor accuracy. Our experiments reiterate these findings. We also show how all the existing schemes fail catastrophically when there is no power law in the labels.   
With current strategies, the cost of improving the quality with current alternatives does not seem worth it over the GPU acceleration of softmax. 


\textbf{Samplers Based on Probabilistic Hash Tables:} In this paper, we change this. Our work provides two families of truly (near) constant O(1) time adaptive sampling schemes utilizing the recent advances in Locality Sensitive Sampling \citep{spring2017scalable, spring2017new,beidi_sgd,charikar2017hashing,luo2019scaling,spring2020mutual}, and exploits the data structure proposed in \citep{daghaghi2021accelerating, chen2019slide}. We provide an efficient implementation of our proposal on CPU, which
outperforms TensorFlow's implementation of softmax and negative sampling strategies on some of the best available GPUs (V100) in terms of wall-clock training time.   

\textbf{Summary of Contributions:}

1) We propose \textbf{two} efficient schemes for negative sampling where the negative sampling distribution provably adapts to changing parameters and the data instance. Furthermore, the sampling cost is provably constant (independent of the number of classes)

2) We show that our technique is not only provably adaptive but also practical. We provide an efficient CPU implementation, in C++, of our negative sampling approach \footnote{The code available at https://github.com/RUSH-LAB/SLIDE}. We demonstrate the effectiveness of a truly (near) constant time negative sampler by showing that our C++ CPU implementations significantly outperform several popular TensorFlow alternatives in wall-clock speed, even when the baselines leverage the powerful V100 GPU Acceleration. In addition, our principled proposed negative sampling schemes achieve the highest accuracy compared to popular heuristics. 

3) We provide a rigorous evaluation of our proposal with its efficient implementation against full softmax and popular approximations like sampled softmax, frequency-based sampled softmax, top-K activation softmax, and Noise Contrastive Estimation (NCE). We report the time-wise and iteration-wise precision on large datasets like Amazon-670K, Wiki-325K, Amazon-Uniform, and ODP-105K. 

\subsection{LSH Based Hash Tables}\label{sec:lsh}
In this section, we briefly describe the recent development of using locality sensitive hashing for sampling and estimation~\citep{spring2017scalable, spring2017new,beidi_sgd,charikar2017hashing,luo2019scaling,spring2020mutual}. Locality Sensitive Hashing \citep{indyk1998approximate,indyk2006polylogarithmic} is a widely used paradigm for large scale similarity search and nearest neighbor search. LSH is a family of hash functions with a unique property that vectors `close' \textit{wrt} some distance metric are more likely to have the same hash code as opposed to vectors that are `far' from each other. Formally, one sufficient condition for a hash family $\mathcal{H}$ to be an LSH family is that the \emph{collision probability} ${Pr}_\mathcal{H}(h(x) = h(y))$ is a monotonically increasing function of the similarity: 
\begin{equation}\label{eq:monotonic}
\small
Pr_\mathcal{H}(h(x) = h(y)) = f(Sim(x,y)),
\end{equation} 
where $f$ is a monotonically increasing function.



The idea is to use the hash value of $x$, i.e., $h(x)$, to generate key of $x$ in the hash table. We first initialize $L$ hash tables by constructing a meta-LSH hash function using $K$ independent hash functions for each of them. For details, see~\citep{andonie2lsh}. There are three major steps:

{\bf Pre-processing Phase:} Given a dataset of size $n$, we first insert all the data points into the hash tables using the meta-LSH formed by concatenating $K$ independent LSH hash functions. We only store the index/pointer of the data point in the hash tables instead of the entire vector. The cost of the addition is $K \times L$ hash computations followed by $L$ insertions in the buckets. 

{\bf Query Phase:} During the query phase, we use the same meta-LSH hash to compute the hash codes for the query. Then we probe the corresponding bucket of each table and retrieve samples from it. The union of candidates from all hash tables constitutes the samples for the particular query. 

{\bf Update Phase:} If an existing element in the database is updated, we can delete the existing element from the hash table and add the updated one. The cost is equivalent to twice the insertion cost of an element which is $2\times K \times L$. 
\begin{figure}{H}
    \centering
    \includegraphics[width = 0.5\textwidth]{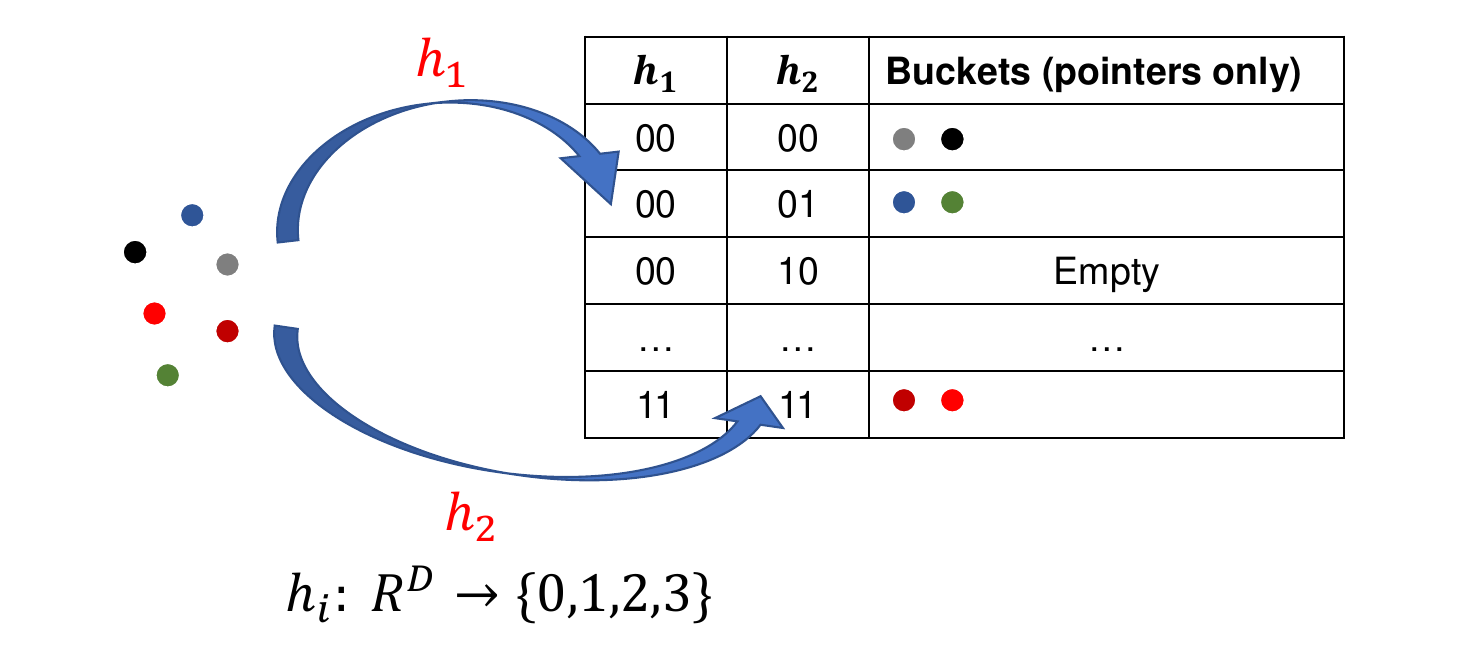}
    \caption{Schematic diagram of LSH. For an input, we compute hash codes and retrieve candidates from the corresponding buckets.\\
}
    \label{fig:lsh}

\end{figure}
\subsection{Adaptive Sampling view of LSH}
Denote $p_{qx}$ be the probability of retrieving $x$ from the datasets, when queried with a given query $q$. \citep{spring2017scalable,spring2017new} for the first time observed that for $(K, L)$ parametrized LSH algorithm the precise form of $p_{qx} = 1-(1-\alpha^K)^L$ can be  used of adaptive sampling and importance estimation. Here $\alpha$ is the collision probability of query $q$ and $x$ under the given LSH function, i.e. $\alpha = Pr_\mathcal{H}(h(x) = h(q))$;  $p_{qx}$ is monotonic in $\alpha$ which is further monotonic in the similarity between query $q$ and the data element $x$. The similarity measure is dependent on the LSH function in use.


\textbf{Constant Time Sampling:} It should be noted that the cost of sampling is the cost of querying, which is only $K\times L$, for all $K$ and $L$, which holds even for $K=1$ and $L=1$. This sampling cost is independent of the number of elements in the data. Clearly, the probability $p_{qx}$ is dependent on the query, and every element $x$ in the data has a different sampling probability. Thus, even though our sampling scheme induces $n$ different sampling probabilities every time the query $q$ is changed, the sampling cost is independent of $n$, and in fact, is constant.
All this is assuming one $\mathcal{O}(n)$ time preprocessing. 

Since 2016, this efficient sampling view of LSH has been used in a wide range of applications, such as deep neural networks~\citep{spring2017scalable, chen2019slide,luo2019scaling,spring2020mutual}, kernel density estimation~\citep{coleman2020sub,coleman2019sub,charikar2017hashing}, record linkage~\citep{chen2018unique}, and optimization~\citep{chen2018lsh}. 
Recent advances in fast inner product search using asymmetric LSH have made it possible to sample large inner products~\citep{shrivastava2014asymmetric}. Effectively, given a query $q$, it is possible to sample an element $x$ from the database with probability proportional to a monotonic function of inner product $f(q^Tx)$; where $f$ is a monotonically increasing function.

\section{Our Proposal: Locality sensitive Negative Sampling (LNS)} 


\begin{figure*}[ht]
\begin{center}
\centerline{\includegraphics[width=5.5in]{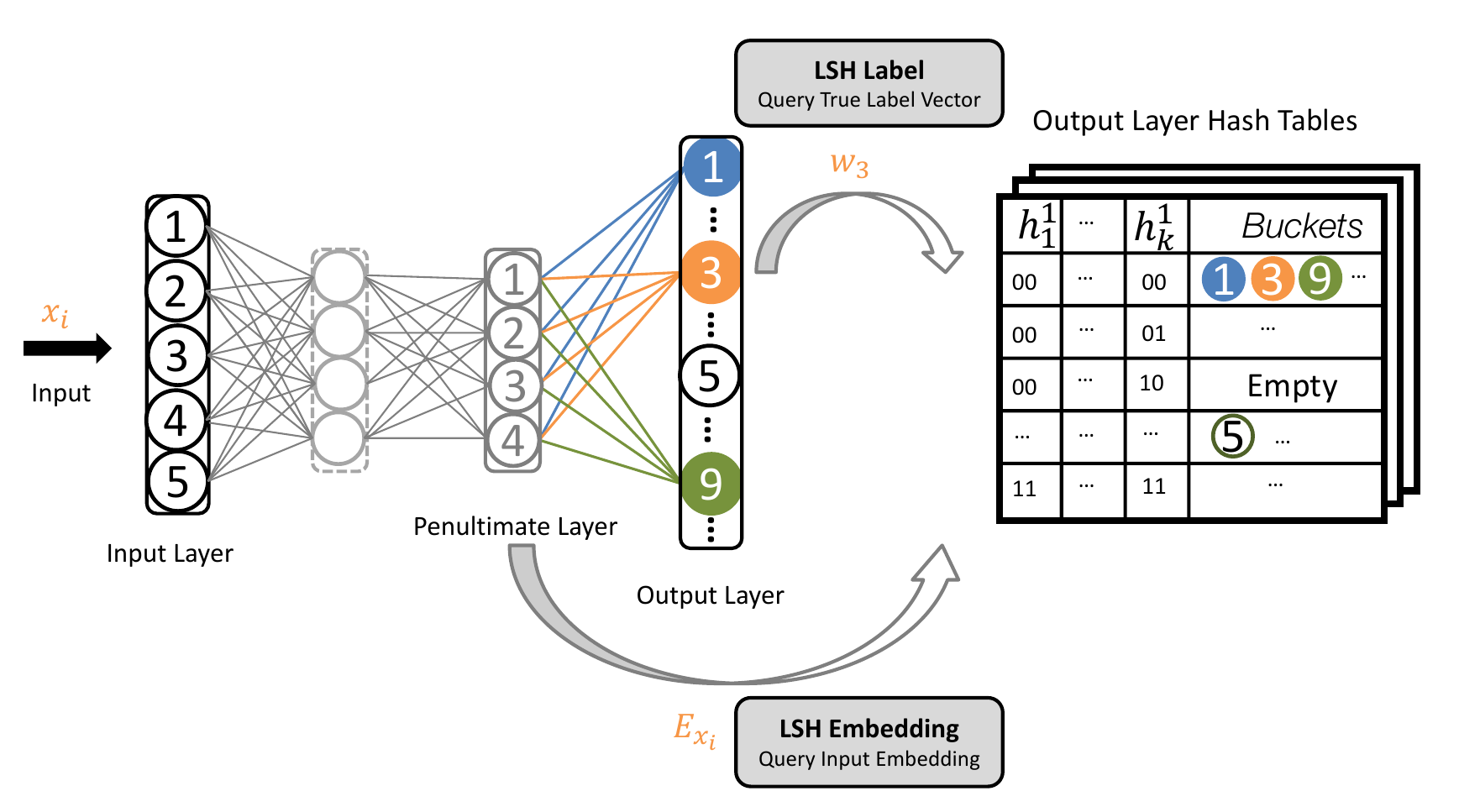}}
\caption{Schematic diagram of our proposal for LSH Label and LSH Embedding schemes. 1) We first construct hash tables for the label vectors $w_i$. The label vectors are the weights of the connections from a label to the penultimate layer. In the figure, e.g. label vector $w_3$ for node 3 (orange node) is the concatenation of its connection weights to the penultimate layer (orange lines). 2) For a training sample $x_i$, we query the LSH tables whether with the true label weights $w_3$ (orange lines) for the LSH Label method, or with the input embedding $E_{x_i}$ for the LSH Embedding method and obtain negative samples (blue and green nodes). We call the retrieved samples `hard' negatives because they are very similar to the `true' ones but are supposed to be `false'.}
\label{fig:Workflow}
\end{center}
\vspace{-5mm}
\end{figure*}

\paragraph{Notations:} We will start by defining a few vectors in the neural network setting illustrated in Figure~\ref{fig:Workflow}. We are in large softmax settings. Here, we will use $N$ to denote the total number of classes. We will define vector $w_i \in \mathbb{R}^d$ (\emph{class vectors}) to be the weight vector associated with class $i$ in the last layer of the neural network. We will use $(x,y)$ to denote the current input sample to the neural network for which we want to generate negative samples. We will use $E_x \in \mathbb{R}^d$ (\emph{final input embedding}) to denote the vector of activation in the penultimate layer of the neural network when fed with input $x$. 

We first describe our sampling procedure, and later we argue why it is distribution-aware and constant time.  Our approach, just like the LSH algorithm, has three phases. The first phase is a one-time costly ($\mathcal{O}(N)$) prepossessing stage. The other two phases, the sampling and update phase, are performed in each iteration, and both of them are constant-time operations independent of $N$. 

{\bf One-time Preprocessing Phase during Initialization:} We start with randomly initializing the neural network parameters. This automatically initializes all the class vectors $w_i$. We now preprocess all these randomly initialized class vectors in $(K, L)$ parameterized LSH hash tables, as described in Section~\ref{sec:lsh}. This is a one-time operation during initialization. 

{\bf Two Negative Sampling Schemes for a given input $(x,y)$:} In this phase, we process input $x$ to the penultimate layer and get the final input embedding $E_x$. Now instead of processing all the $N$ nodes in the last layer, we query the hash tables with either vector $E_x$ (\emph{{\bf LSH Embedding}}) or with the weight vector of the true label $y$, i.e., $w_y$ (\emph{{\bf LSH Label}}). This preciously describes our two sampling schemes. We can obviously mix and match, but we consider these two choices as two different methods for the simplicity of analysis and evaluations.  

When we query, we generate a small set of the sampled candidates, call them $ C $, forming our negative samples. Thus, we only compute the activation of nodes belonging to $C \cup y$ in the last layer and treat others as zero activation.   

{\bf Update Hash Tables with Update in Weights:} During backpropagation for input $(x,y)$, we only update $C \cup y$ weights in the last layer. We update these changed weights in the LSH hash tables.

Next,  we first argue why this sampling is distribution aware and adaptive with every parameter and input change. We will then argue that the sampling and update process is significantly efficient. It is a constant-time operation that is easily parallelizable.

\subsection{What is the Sampling Distribution? Is it Adaptive? Is it Constant Time?}

\begin{definition}
\textbf{Adaptive Negative Sampling}: We call a negative sampling distribution adaptive if the distribution changes with the change in the parameter of the network as well as the change in the input. Essentially, the probability of selecting a class $Pr(y)$ is a non-trivial function of the input $x_i$ and the parameters $W$ of the neural network.   
\end{definition}

\textbf{Comment 1}: Static-based sampling approaches such as sampled softmax \citep{bengio2008adaptive} are not adaptive, since they consider a fixed underlying sampling distribution regardless of the change in the input and the network parameters.

\textbf{Comment 2}: \citet{vijayanarasimhan2014deep} and other variants of LSH utilizes LSH as a subroutine for \emph{top-k search}, which is significantly expensive from both time and memory perspective (requires $N^\rho$ resources, which is too much per iterations). The main realization of our work is that we use LSH for \textit{sampling} which can be even constant time and work on any budget. Please note that LSH for exact \textit{search} is prohibitively expensive in every iteration, while the \textit{sampling} perspective of LSH is super efficient. LSH as search (the standard algorithm) where instead of just sampling from buckets, we retrieve all elements from buckets as candidates. We then filter the candidates to find the top-k (the standard LSH procedure mentioned in \citet{vijayanarasimhan2014deep}).  The per iteration cost of this process for Amazon-670K is 100x slower \citep{chen2019slide} than our sampling process where we just hash, and sample from the bucket.

We start with two theorems that give the precise probability distribution of sampling a class as a negative sample with  LSH Label and  LSH Embedding methods provided the input $(x,y)$ and current parameters. We will use $p_{xy}$ as the collision probability of the LSH hash value of $x$ and $y$.

\begin{theorem}
{\bf LSH Label Distribution} For an input $(x,y)$ and LSH parameters $(K,L)$, the probability of sampling a class $i \ne y$ as negative sampling with {\bf LSH Label}  method is given by
$$p_i \propto 1 -(1- p_{w_yw_i}^K)^L,$$ where $w_y$ and $w_i$ are the weights associated with true class $y$ and class $i$ respectively. Furthermore, the probability of sampling class $i$ is more than any other class $j$, if and only if $sim(w_y,w_i) > sim(w_y,w_j)$. Here $sim$ is the underlying similarity function of the LSH.
\end{theorem}

\begin{theorem}
{\bf LSH Embedding Distribution} For an input $(x,y)$ and LSH parameters $(K,L)$, the probability of sampling a class $i \ne y$ as negative sampling with {\bf LSH Embedding} method is given by $$p_i \propto 1 -(1- p_{E_xw_i}^K)^L,$$ where $E_x$ is  the embedding vector of input x and $w_i$ is the weights associated with class $i$ respectively. Furthermore, the probability of sampling class $i$ is more than any other class $j$, if and only if $sim(E_x,w_i) > sim(E_x,w_j)$. Here $sim$ is the underlying similarity function of the LSH.
\end{theorem}

\vspace{-2mm}
\textbf{Comments:} The expressions of probability are immediate from the sampling view of LSH. The expressions $1 - (1 - p^K)^L$ is monotonically increasing in $p$, the collision probability, which in turn is monotonically increasing in the underlying similarity function $sim$. Clearly, the distribution is adaptive as they change with the input $(x,y)$ as well as the parameters. So any update in the parameter or any change in the input changes the sampling distribution completely. However, the sampling cost is constant and independent of the number of classes we are sampling from.

\textbf{Computational Cost for Processing Each Input:}
Given an input $(x,y)$, the cost of processing it without any negative sampling is $\mathcal{O}(N)$. With our proposed negative sampling the cost of sampling is the cost of query which is $K \times L$, a negligible number compared to $N$ in practice.

The cost of the update is slightly more  $(|C|+1) \times K \times L$ because we have to update $|C|+1$ weights. In negative sampling, $C$ is a very small constant. Also, in practice $K$ and $L$ are constants. Furthermore, we have a choice to delay the hash table updates. 

\paragraph{Intuition of LSH Label:} Coming back to our example of class 'Nike Running Shoes'. Let us focus on LSH Label distribution. Initially, when all other labels have random weights, the similarity between the label 'Nike Running Shoes' and any other label will be random. So initial negative sampling should be like uniform sampling.  However, as the learning progresses, it is likely that 'Nike Running Shoes' and `Adidas Running Shoes' will likely get close enough. Their weights will have high similarity (high $sim$), at that time, the LSH Label sampling will select `Adidas Running Shoes' as a likely negative sample for `Nike Running Shoes' class. 
\vspace{-2mm}
\paragraph{Intuition of LSH Embedding:} The LSH Embedding method is also adaptive. Consider the similarity function as an inner product. Input embedding inner product with class vector is directly proportional to its activation. Thus, it naturally selects classes in which the classifier is confused (high activation but incorrect) as negative samples. Again, the distribution is adaptive.  


\begin{figure} [ht]
	\begin{center}
	\includegraphics[width=3in]{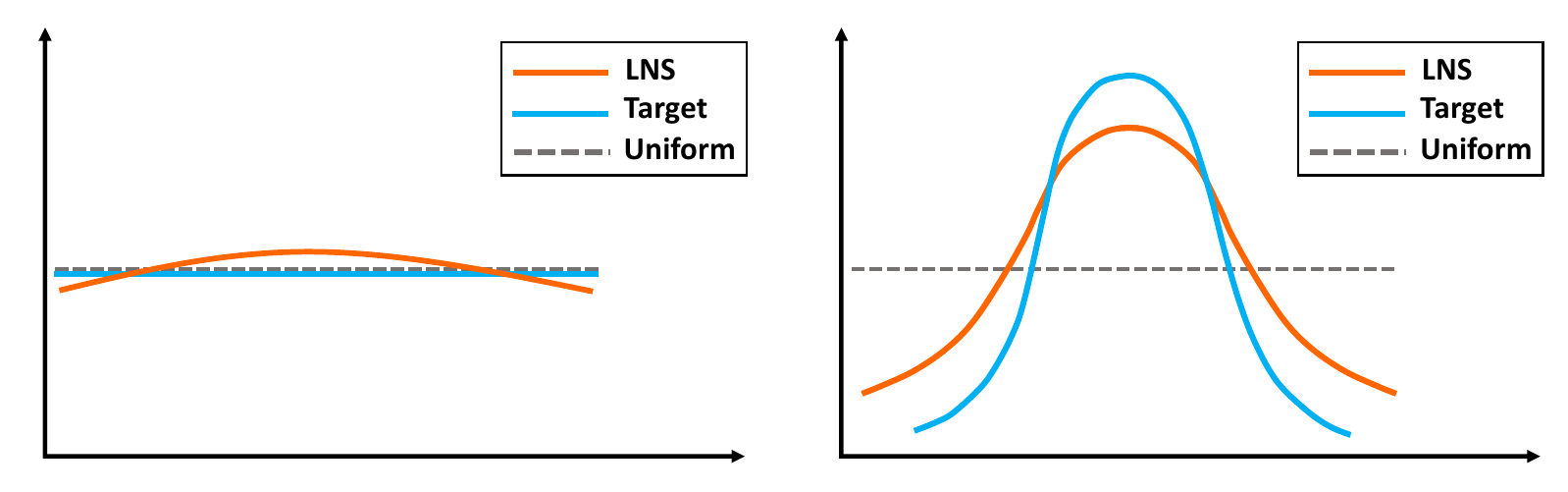}
	\end{center}
	\caption{How the true negative sampling distribution (target), uniform negative sampling and LNS adapts over iterations. Initially, when there is no learning, the sampling distribution is close to uniform (\emph{left figure}). During later states the sampling distribution is significantly different from uniform (\emph{right figure}). The LNS is adaptive and distribution-aware and it follows the true distribution. }
	\label{fig:distribution}
\end{figure}

\subsection{Algorithm and Implementation Details}
First, we construct $K \times L$ hash functions and initialize the weights of the network and $L$ hash tables. The LSH hash codes of weight vectors of the last layer are computed and the id of the corresponding neuron is saved into the hash buckets (Algorithm \ref{alg: alg_prepro}). During the feed-forward path in the last layer, we query whether the embedding vector (LSH Embedding scheme) or the label vector of true class (LSH Label scheme) and retrieve the classes from hash table which are considered as negative classes. Instead of computing the activation of all the output nodes (full softmax), we compute the activations of the true classes and the retrieved negative classes. For the backpropagation, we backpropagate the errors to calculate the gradient and update the weights for the active nodes. Please refer to Algorithm \ref{alg: alg_main}, Algorithm \ref{alg: alg_smpl}, Algorithm \ref{alg: alg_query}, and \ref{alg: alg_update} for more details.  
\begin{algorithm}[!htb]
\caption{Locality Sensitive Negative Sampling (LNS)}
\begin{algorithmic}[1]
\INPUT Input data $(X,Y)$, $N$ number of classes, $S_p$ sparsity
\OUTPUT $C$ set of active neurons of the last layer

\STATE Initialize weights $W_l$ for the last layer $l$

\STATE {$T, h$ = Preprocessing ($W_l$) (Algorithm 2)}

\FOR{\textit{each iteration}}
\STATE {Batch = $(x,y)$}
\STATE {Compute final input embedding $E_x$ and class vectors $w_i$  in the forward path}
\IF{LSH Embedding}
\STATE{$C$ = Sampling($E_x,T, h, N, S_p$)} (Algorithm 3)
\ENDIF
\IF{LSH Label}
\STATE{$C$ = Sampling($w_i,T, h, N, S_p$)} (Algorithm 3)
\ENDIF
\STATE {Backpropagation($C\cup y$)}
\STATE{$T$ = UpdateHashTables} (T, $W_{i\in C \cup y}^{old,new}$) (Algorithm 5)
\ENDFOR

\label{alg: alg_main}
\end{algorithmic}
\end{algorithm}
\vspace{-5mm}


\begin{algorithm}[!htb]
\caption{Preprocessing}
\begin{algorithmic}[1]
\INPUT Data $\mathcal{D}$ size $n$
\OUTPUT $L$ hash tables, $K \times L$ LSH functions 
\STATE {Create hash tables $T_1, ..., T_L$}
\STATE {Create $K \times L$ LSH functions $h_{k,l}$}
\FOR{$x_i\in \mathcal{D}$}
\STATE {Compute $K \times L$ hash values $h_{k,l}(x_i)$}
\FOR{Hash table $T_t$, $t=1:L$}
\STATE {Concatenate $h_{1,t}(x_i),h_{2,t}(x_i),...,h_{k,t}(x_i)$ to construct the meta-hash value $H_t(x_i)$}
\STATE {Map $H_t(x_i)$ to bucket $b$} 
\STATE {Insert $x_i$ into $T_t(b)$}

\ENDFOR

\ENDFOR
\STATE {\textbf{return} $T, h$}

\label{alg: alg_prepro}
\end{algorithmic}
\end{algorithm}



\begin{algorithm}[!htb]
\caption{Sampling}
\begin{algorithmic}[1]
\INPUT $q$ query, $T$ hash tables, $h_{k,l}$ $K \times L$ LSH functions,  $N$ number of classes, $S_p$ sparsity
\OUTPUT $S$ set of retrieved samples from hash tables 
\STATE {$S = \emptyset$}
\FOR{$t=1:L$}

\IF{$|S|/N$ $\leq$ $S_p$}
\STATE{$S = S$  $\cup$ Query ($q, h_{k,t}|_{k=1}^{k=K}, T_t$)} (Algorithm 4)

\ELSE
\STATE{\textbf{break}}
\ENDIF

\ENDFOR
\STATE {\textbf{return} $S$}

\label{alg: alg_smpl}
\end{algorithmic}
\end{algorithm}



\begin{algorithm}[!htb]
\caption{Query (Negative Sampling on Fly)}
\begin{algorithmic}[1]
\INPUT $q$ query, $h_{k,T}$ as $K$ LSH hash functions, $T$ hash table
\OUTPUT  $S$ retrieved samples
\STATE {Compute query hash values $h_{k,T}(q)|_{k=1}^{k=K}$}

\STATE {Concatenate $h_{1,T}(q),h_{2,T}(q),...,h_{k,T}(q)$ to compute the meta-hash value $H_T(q)$}
\STATE {Map $H_T(q)$ to bucket $b$}
 \STATE {$S =  T(b)$}

\STATE {\textbf{return} $S$}

\label{alg: alg_query}
\end{algorithmic}
\end{algorithm}


\begin{algorithm}[!htb]
\caption{UpdateHashTables}
\begin{algorithmic}[1]
\INPUT T hash tables, $w_i^{old}$, $w_i^{new}$ the old and the updated weight vectors of negative classes $C$ and true classes $y$
\OUTPUT  $T$ updated hash tables 
\FOR{$w_i^{old}$, $i \in C \cup y$}
\STATE{Compute hash values of $w_i^{old}$} (run steps \{5:7\} of Algorithm 2 for $w_i^{old}$)
\STATE{Delete $w_i^{old}$ from hash tables }
\ENDFOR
\FOR{$w_i^{new}$, $i \in C \cup y$}
\STATE{Compute hash values of $w_i^{new}$} (run steps \{5:7\} of Algorithm 2 for $w_i^{new}$)
\STATE{Insert $w_i^{new}$ into hash tables}
\ENDFOR

\STATE {\textbf{return} $T$}

\label{alg: alg_update}

\end{algorithmic}
\end{algorithm}

\section{Experiments}

In this section, we will empirically evaluate the performance of our LSH Negative Sampling (LNS) approach against other sampling schemes that are conducive to GPUs. The real advantage of LNS is noticeable with huge neural networks. The popular extreme classification challenges have models with more than 100 million parameters, which are ideal for our purpose. For these challenges, most of the heavy computations happen in the last layer. 

\subsection{Datasets} 
We evaluate our framework and other baselines on four datasets. Amazon-670K and Wiki-325K are two multi-label datasets from extreme classification repository~\citep{Bhatia16}, ODP is a multi-class dataset which is obtained from \citep{NIPS2015_e369853d},
and Amazon-Uniform is a variant of Amazon-670K dataset with uniform label distribution [\ref{sec: powerlaw}]. The statistics about the dimensions and samples sizes are shown in Table \ref{table:data}, for more details see Section \ref{appendix:dataset} in the Appendix.

\label{sec:dataset}
\begin{table}[H]
\vspace{-0.2in}
    \centering
\tiny
\footnotesize
    \caption{Statistics of the datasets}
    \begin{tabular}{c|c|c|c|c} \hline
    	Dataset & Feature Dim  & Label Dim & \#Train & \#Test \\ \hline
    	Amz-670K & 135909  & 670091 & 490449 & 153025\\ \hline
    	Wiki-325K & 1617899 & 325056 & 1778351 & 587084\\ \hline
    	ODP & 422713 & 105033
 & 1084320 & 493014\\
    	\hline
    	Amz-Unif & 135909 &158114 & 348174 & 111018\\
    	\hline
    	\bottomrule
    	\end{tabular}
    	\label{table:data}
    	\vspace{-2mm}
\end{table}


\begin{figure*}[h!]
\begin{center}
\begin{multicols}{3}
    \includegraphics[width=0.95\linewidth]{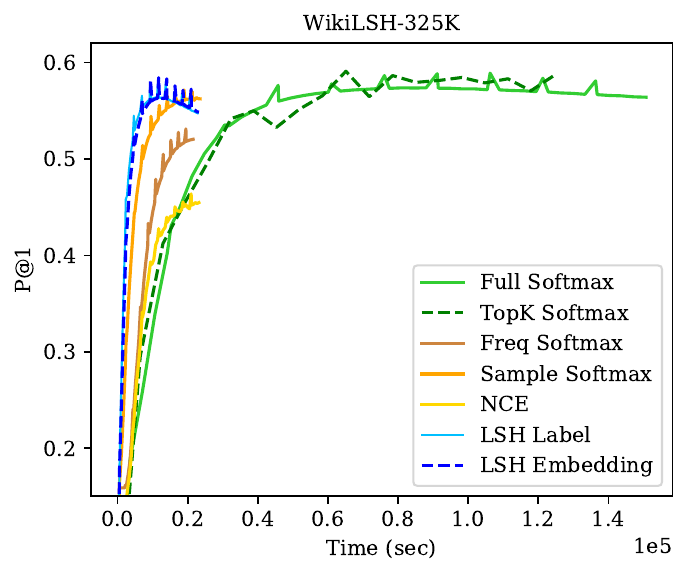}\par 
      \label{}
         \includegraphics[width=0.95\linewidth]{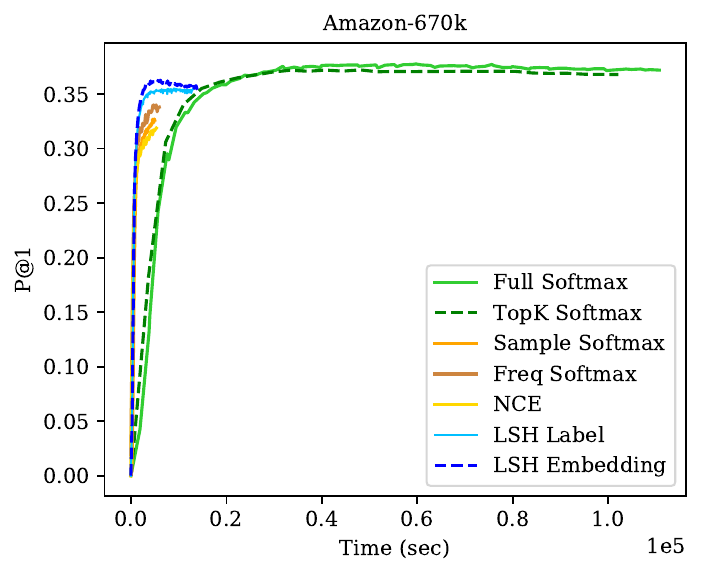}\par 
        \label{} 
            \includegraphics[width=0.95\linewidth]{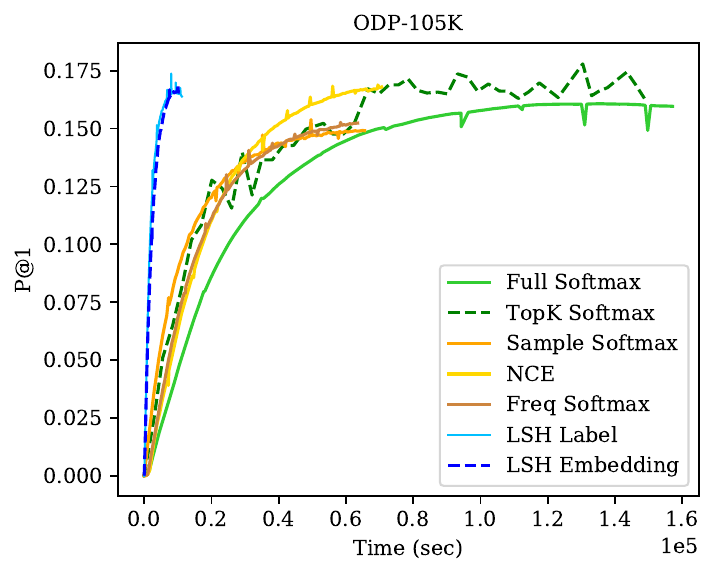}\par
      \label{}
      \label{}
    \end{multicols}
    \begin{multicols}{3}
    \includegraphics[width=\linewidth]{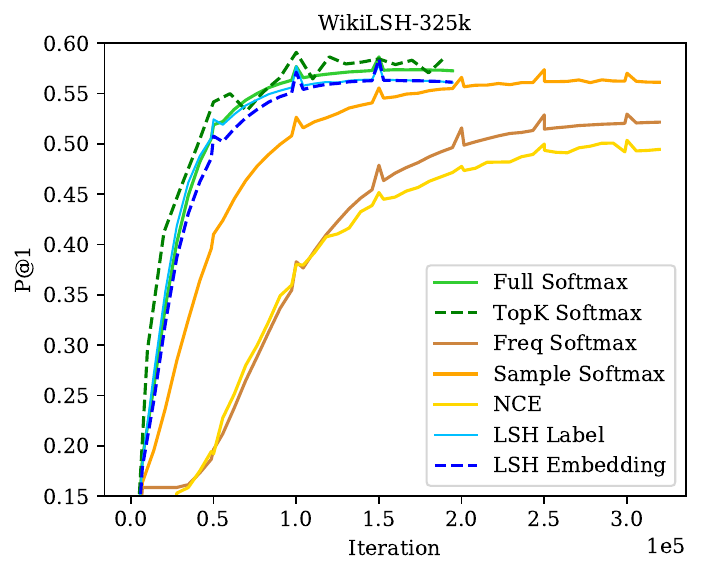}\par 
      \label{}
    \includegraphics[width=\linewidth]{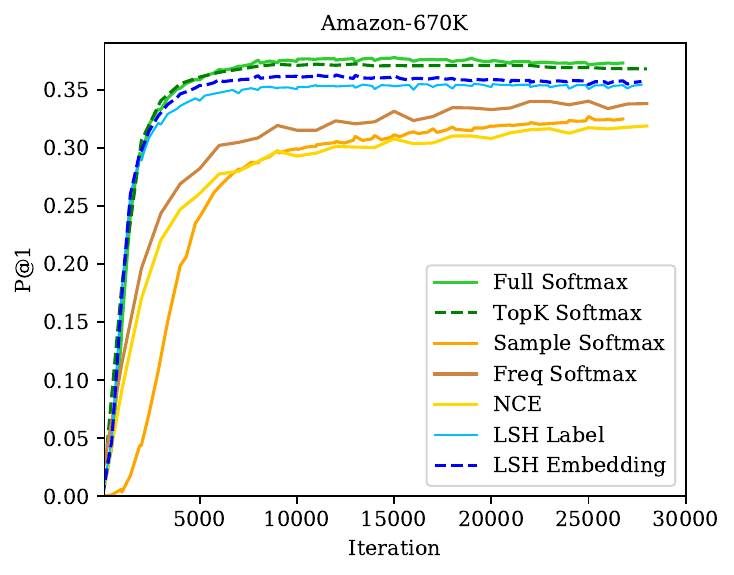}\par 
          \label{}
        \includegraphics[width=\linewidth]{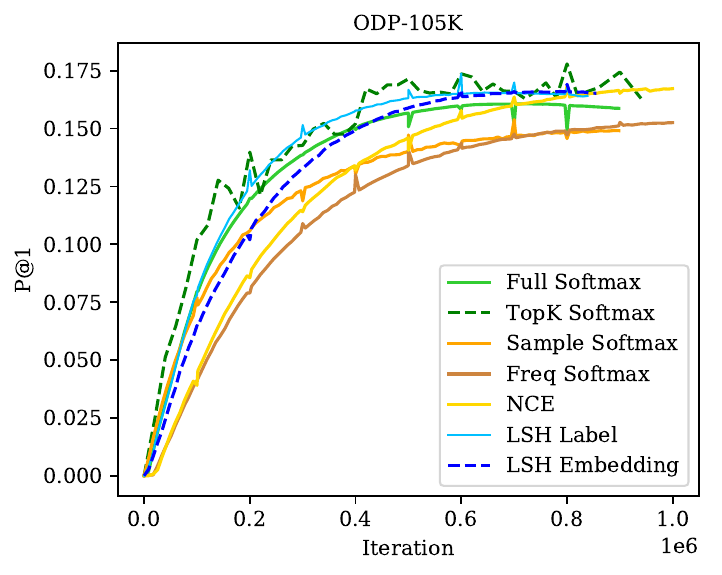}\par
      \label{}
      \label{}
    \end{multicols}
\end{center}
\vspace{-0.2in}
\caption{ Comparison of our proposal LNS with two schemes (LSH label and LSH embedding, both on CPU) against five baselines: Full softmax, TopK softmax, Frequency-based softmax, Sampled softmax and NCE (all on NVIDIA V100 GPU with Tensorflow) for three datasets. {\bf Top Row:} Precision@1 vs time, {\bf Bottom Row:} Precision@1 vs iteration, {\bf Left Column}: Wiki-325K dataset \textbf{Middle Column}: Amazon-670K dataset \textbf{Right Column}: ODP dataset. The time-wise plots (top row) are representative of comparison w.r.t the \emph{average time per epoch} metric. The LSH methods closely mimic Full softmax in iteration-wise plots indicating the superiority of distribution-aware sampling. The time plots clearly indicate the speed of sampling, where LSH samplings are the best-performing ones.} 
\vspace{-1mm}
\label{fig:main_plot}
\end{figure*}

\vspace{-3mm}
\subsection{Baselines} We benchmark our proposed framework against Full softmax, Sampled softmax, TopK softmax, Frequency-based softmax and Noise Contrastive Estimation (all explained below). All the baselines use TensorFlow and run over NVIDIA V100 GPU. To have a fair comparison, the architecture, optimizer, and size of hidden layer are exactly the same for all the methods on each dataset. Please note that our proposed schemes are implemented in C++ and experiments are performed over CPU. Despite this hardware disadvantage, they still outperform the other methods due to the efficiency of the process.

 \textbf{Full Softmax}: Full softmax updates the weights of all the output neurons, which makes it computationally expensive and intractable for extreme classification framework.\\
 \textbf{Sampled Softmax}: Sampled softmax draws negative samples based on log-uniform distribution and updates their corresponding weights plus the weights for the true classes. This approach alleviates the computational bottleneck but degrades the performance in terms of accuracy.\\
\textbf{TopK Softmax}: TopK softmax updates the weights of the output neurons with $k$ highest activations (including the true classes). This framework maintains better accuracy than Sampled softmax but with a slower convergence rate due to the scoring and sorting of all activations. \\
\textbf{Frequency based Softmax:}
Frequency-based softmax samples the classes in proportion to the frequency of their occurrence in the training data. Computationally, this is the same as Sampled softmax, however, it samples negative classes from more frequent classes with higher probability.\\
\textbf{Noise Contrastive Estimation (NCE):} NCE loss~\citep{pmlr-v9-gutmann10a} tackles multi-class classification problem as multiple binary classifiers instead. Each binary classifier is trained by logistic loss to distinguish between true classes and negative classes. Negative classes are sampled from a noise distribution which is typically log-uniform distribution or based on class frequencies. 
\\
\textbf{LNS (our proposal)}: Our proposed negative sampling algorithm samples the classes from output distribution which is adaptive to the input, true class, and model parameters. Our model utilizes LSH to sample the most confusing (the most similar but false) classes as the negative samples in (near) constant time. We implement and compare both {\bf LSH Label} and {\bf LSH Embedding} schemes. 






\subsection{Architecture and Hyperparameters} 


We use a standard fully connected neural network with a hidden layer size of 128 for all datasets,
and we performed hyperparameter tuning for all the baselines to maintain their best trade-off between convergence time and accuracy. The optimizer is Adam with a learning rate of 0.0001 for all the experiments. The batch size for Amazon-670K, Wiki-325K, Amazon-Uniform, and ODP is 1024, 256, 256, and 128 respectively for all the experiments. We apply hash functions for the last layer where we have the computational bottleneck. In LSH literature, $L$ denotes the number of hash tables and $K$ denotes the number of bits in the hash code for each hash table (thereby having $2^K$ buckets per hash table).
We use DWTA hash function (see section \ref{appendix: Dwta} for details) for all datasets, with K=5 and L=300 for Wiki-325K, K=6 and L=400 for Amazon-670K, K=5 and L=150 for ODP, and K=6 and L=150 for Amazon-Uniform. We update the hash tables with an initial update period of 50 iterations and then exponentially decaying the updating frequency (as we need fewer updates near convergence). Our experiments are performed on a single machine with 28-core and 224-thread processors. All the baselines are run on the state-of-the-art NVIDIA V100 GPUs with 32 GB memory.

\begin{table*}[ht!]
\vspace{-0.1in}

\begin{center}

  \caption{Comparison of LSH Embedding and LSH Label against the other baselines w.r.t the epoch  number that P@1 reaches 50\% of Full softmax final P@1, the epoch  number that P@1 reaches 90\% of Full softmax final P@1, and \emph{precision@1} (P@1). $Ei$ means that at epoch $i$ the method reaches 90\% of Full softmax P@1, and \emph{'Fail'} means that the method fails to reach 90\% of Full softmax P@1. Our proposals, LSH Embedding and LSH Label, run on CPU, while all the five baselines run on NVIDIA V100 GPU with Tensorflow.   }


\scriptsize
  \begin{tabular}{p{1.35cm}|p{1.1cm}p{1.1cm}p{0.4cm}|p{1.1cm}p{1.1cm}p{0.4cm}|p{1.1cm}p{1.1cm}p{0.4cm}|p{1.1cm}p{1.1cm}p{0.4cm}} 

    \toprule
   \hline
        \multicolumn{1}{c}{}
      &\multicolumn{3}{c}{Amazon-670K}
    &\multicolumn{3}{c}{Wiki-325K}
    &\multicolumn{3}{c}{ODP-105K}
    &\multicolumn{3}{c}{Amaz-Uniform}

    \\
    
    Method &\#epochs to reach 50\% of Acc  &\#epochs to reach 90\% of Acc  &P@1  &\#epochs to reach 50\% of Acc  &\#epochs to reach 90\% of Acc    &P@1  &\#epochs to reach 50\% of Acc  &\#epochs to reach 90\% of Acc    &P@1 &\#epochs to reach 50\% of Acc   &\#epochs to reach 90\% of Acc   &P@1   \\
    \hline
   Full Soft   &E2  &E6 &37.5 &E2  &E7 &57.3  &E12 &E40  & 16.2  &E2 &E2  &24 
\\
    \hline
     LSH Embed   &E2  &E6   &36.1  &E2 &E9  &56.3 &E16 &E44  &16.8   &E2  &E4   &22.8 
    \\
           \hline
        LSH Label &E2  &E8 &35.5 &E2 &E8  &56.1 &E12 &E35   &16.7  &E2  &E5  &22.5 \\
        \hline
       TopK Soft  &E2 &E5 &37.2  &E1 &E7 &57.5  &E11 &E34  &17.2 &E2   &E6  & 24 
    \\
    \hline
     
     Freq Soft  &E5 &E41 &34  &E11 &E31  & 52.1 &E24  &E82 & 15.2  &E22   &\emph{Fail}  &19.2  \\
     \hline
     
     Sampled Soft &E8  &\emph{Fail}   &32.4  &E4 &E15  & 55.7    &E14 &E78    &14.8  &E8 &\emph{Fail} &20.7 \\
     \hline
     
     NCE  &E5 &\emph{Fail}  & 31.8 &E11 &\emph{Fail}   & 49.9   &E21 &E59   &17  &E32  &\emph{Fail} &16.1    \\
     \hline
    \bottomrule

\end{tabular}
   \label{table:table_epoch}
\end{center}

\end{table*}

\begin{table*}[ht]
\vspace{-0.1in}

\begin{center}

  \caption{Comparison of LSH Embedding and LSH Label against the other baselines w.r.t the Average training time per epoch, \emph{precision@1} (P@1) (\%) and \emph{precision@5} (P@5)(\%). Our proposals, LSH Embedding and LSH Lable, run on CPU, while all the five baselines run on NVIDIA GPU with Tensorflow. This table represents \emph{average training time per epoch} metric as opposed to the \emph{convergence time} metric. }
\vskip 0.05in


\scriptsize
  \begin{tabular}{p{1.35cm}|p{1.3cm}p{0.4cm}p{0.4cm}|p{1.3cm}p{0.4cm}p{0.4cm}|p{1.3cm}p{0.4cm}p{0.4cm}|p{1.3cm}p{0.4cm}p{0.4cm}} 

    \toprule
  \hline
        \multicolumn{1}{c}{}
      &\multicolumn{3}{c}{Amazon-670K}
    &\multicolumn{3}{c}{Wiki-325K}
    &\multicolumn{3}{c}{ODP-105K}
    &\multicolumn{3}{c}{Amaz-Uniform}

    \\
    
    Method &Avg training time per epoch &P@1 &P@5 &Avg training time per epoch  &P@1 &P@5 &Avg training time per epoch   &P@1 &P@5 &Avg training time per epoch &P@1 &P@5  \\
    \hline
  Full Soft   & Baseline  &37.5 &33.6 & Baseline  &57.3 &51.7 & Baseline & 16.2 &29.2 & Baseline   &24 &32.5
\\
    \hline
     LSH Embed    & 11x  &36.1 &33.5  & 6.5x   &56.3 &46  &15x   &16.8 &32.2    &22x   &22.8 &33.6 
    \\
          \hline
        LSH Label   &10.3x  &35.5 &33.2 &6.6x   &56.1 &46.3 & 14x  &16.7 &31.7  	& 22.6x  &22.5 &33.2\\
        \hline
      TopK Soft  & 1.1x   &37.2 &33.5  & 1.18x &57.5 &52.1 &1.2x   &17.2 &32.4 &1.23x   & 24 &32.5 
    \\
    \hline
     
     Freq Soft & 18.4x   &34 &29.5 & 6.5x  & 52.1 &45.6 & 2.5x  & 15.2 &28.8  & 6x  &19.2 &30.9 \\
     \hline
     
     Sampled Soft  &21x     &32.4 &30.2  &7.8x & 55.7  &48.3  &2.7x  &14.8 &29.1 &6.3x &20.7 &30.1 \\
     \hline
     
     NCE  &20.5x  & 31.8 &29.3 &7x  & 49.9 &41.5 &2.6x  &17  &32  & 6.15x  &16.1  &20.7   \\
     \hline
    \bottomrule

\end{tabular}
  \label{table:table_time}
\end{center}
 \vspace{-0.25in}

\end{table*}

\vspace{-2mm}
\subsection{Results}
We provide numerical results in terms if two metrics. Table \ref{table:table_time} shows the comparisons in terms of \emph{average training time per epoch}, and Table \ref{table:table_epoch} shows the comparisons in terms of \emph{convergence epoch}, i.e the epoch number the model reaches 90\% and 50\% of Full softmax final accuracy.  
 Figure \ref{fig:main_plot} shows the plots comparing $Precision@1$ (denoted here by P@1) versus both wall-clock training time and the number of iterations for our method and all the baselines. 
 For Amazon-670K dataset, LSH Label and LSH Embedding are respectively \textbf{10.3x} and \textbf{11x} faster than TensorFlow Full softmax on GPU in terms of \emph{average training time per epoch} while maintaining the accuracy. Note that although Sampled softmax and NCE are faster than our proposal in terms of \emph{average training time per epoch}, it takes them 8 and 5 epochs, respectively, to reach 50\% of Full softmax accuracy, while it takes only 2 epochs for LSH Embedding and LSH Label. Moreover, Sampled Softmax and NCE fail to reach 90\% of Full softmax accuracy, while it takes only 6 and 8 epochs for LSH sampling methods to reach this level of accuracy.
The same is true for Wiki-325K dataset where LSH Label and LSH Embedding are \textbf{6.5x} faster than TensorFlow Full softmax on GPU, while Sampled softmax and NCE speed up w.r.t. \emph{average training time per epoch} is negligible compared to their \emph{convergence time} and their low accuracy. For the ODP dataset, our proposal significantly outperforms the other baselines in terms of time and accuracy where LSH Label and LSH Embedding achieve \textbf{14x} and \textbf{15x} speed up over Full softmax, and preserve the accuracy. Although NCE achieves competitive accuracy, it is around 5.6x slower than our algorithm in terms of \emph{average training time per epoch}, also it converges slower than our algorithm in terms of \emph{convergence epoch}.  Similarly, for the Amazon-Uniform dataset, LSH Embedding and LSH Label outperform all the other baselines with a significant margin. Our proposal achieves more than \textbf{22x} speedup over Full softmax on GPU in temrs of \emph{average training time per epoch}, while maintains accuracy. See Section \ref{sec: powerlaw} for Amazon-Uniform results. 

Clearly, both variations of our LNS method outperform other negative sampling baselines on all datasets. Static negative sampling schemes, although fast per epoch wise, fail to reach good accuracy. The accuracy climb is also slower due to the poor negative sampling. Our proposal even after drastic sub-sampling is very similar to Full softmax iteration-wise. The results establish the earlier statement that LNS does not compromise performance for speed-up. This is particularly noteworthy because our implementation of LNS uses only CPU while all other baselines run on NVIDIA V100 GPU with TensorFlow. See supplementary material for more experiments.

\subsection{Non-Power Law Label Distribution}
\label{sec: powerlaw}
\begin{figure}[H]
\vspace{-0.1in}
\begin{center}
\begin{multicols}{2}
    \includegraphics[width=\linewidth]{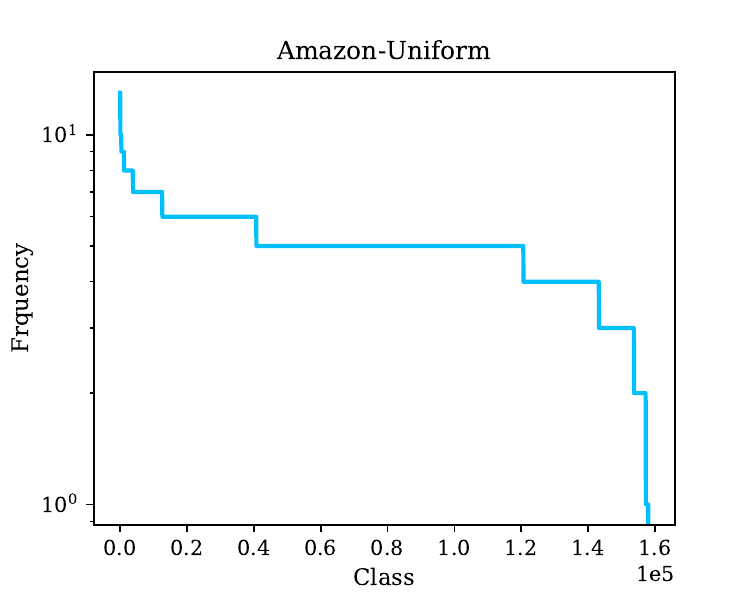}\par 
      \label{}
        \includegraphics[width=\linewidth]{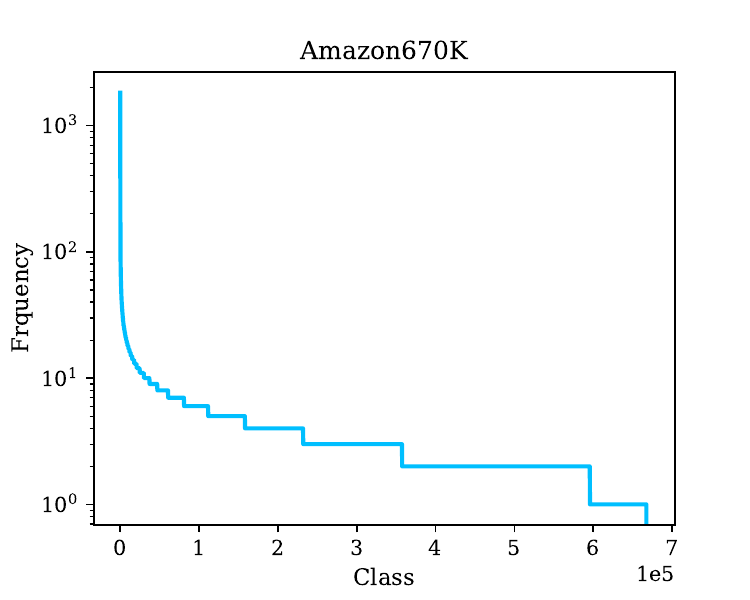}\par
      \label{}
    \end{multicols}
    \begin{multicols}{2}
    \includegraphics[width=\linewidth]{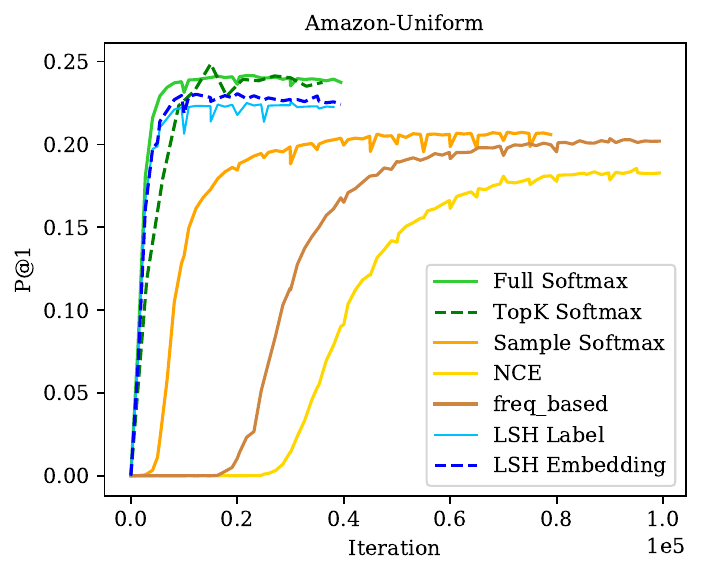}\par 
      \label{}
    \includegraphics[width=\linewidth]{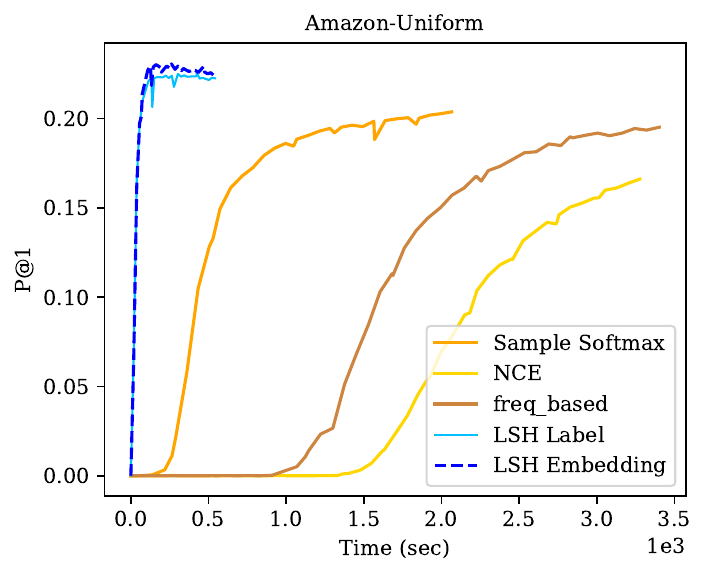}\par
      \label{}
    \end{multicols}
\end{center}
\vspace{-4mm}
\caption{\textbf{Top Row:} Label frequency for Amazon-Uniform (\textit{left column}) and Amazon-670K datasets (\textit{right column}). The label distribution for Amazon-Uniform is near-uniform and it does not follow power law as opposed to Amazon-670K. \textbf{Bottom Row:} P@1 w.r.t iteration (\emph{left figure}) and wall-clock training time (\emph{right figure}) for Amazon-Uniform. LSH label and LSH Embedding outperform NCE and Sampled Softmax by a significant margin.}
\label{fig:label_dist}
\vspace{-0.1in}
\end{figure}

Class distribution in most public available datasets follows the power law, i.e. distribution is long tailed and dominated by high frequent classes. That is why sampling methods like Sampled softmax and NCE, with fixed underlying log-uniform distribution, have acceptable performance on these datasets. To highlight the effectiveness and generality of our proposal method against popular Sampled softmax and NCE on datasets with non-power law labels, we create a variant of Amazon-670K dataset with uniform label distribution by down sampling frequent classes. The new dataset, called Amazon-uniform, has 158K classes and its label distribution is near uniform. The top row in Figure \ref{fig:label_dist} denotes the label distribution of the new dataset against Amazon-670K, which is clearly near uniform. The bottom row in Figure \ref{fig:label_dist} includes convergence plots with respect to the time and iteration. Full softmax and TopK are not included in the time-wise plot for a better representation. Please refer to Table \ref{table:table_epoch} and Table \ref{table:table_time} for the details on these baselines.
The plots confirm the failure of Sampled softmax and NCE, since their underlying sampling distribution is log-uniform and based on the power law assumption. However, our proposed method achieves more than \textbf{22.5x} speed up over Full softmax, and highly outperforms Sampled softmax and NCE with respect to time and accuracy. Our algorithm is truly adaptive and distribution-aware regardless of the label distribution. 

\vspace{-6mm}
\section{Conclusion}
We proposed two efficient and adaptive negative sampling schemes for neural networks with an extremely large number of output nodes. To the best of our knowledge, our proposed algorithm is the first negative sampling method that samples negative classes in near-constant time, while adapts to the continuous change of the input, true class, and network parameters. We efficiently implemented our algorithm on CPU in C++ and benchmarked it against standard TensorFlow implementation of five baselines on GPU. Our method on CPU outperforms all the TensorFlow baselines on NVIDIA GPU with a significant margin on four datasets. 

\vspace{-0.1in}
\section{Acknowledgment}
This work was supported by National Science FoundationIIS-1652131, BIGDATA-1838177, AFOSR-YIP FA9550-18-1-0152, ONR DURIP Grant, and the ONR BRC grant on Randomized Numerical Linear Algebra.

\nocite{langley00}

\bibliography{example_paper}

\begin{thebibliography}{37}
\providecommand{\natexlab}[1]{#1}
\providecommand{\url}[1]{\texttt{#1}}
\expandafter\ifx\csname urlstyle\endcsname\relax
  \providecommand{\doi}[1]{doi: #1}\else
  \providecommand{\doi}{doi: \begingroup \urlstyle{rm}\Url}\fi

\bibitem[Andoni \& Indyk(2004)Andoni and Indyk]{andonie2lsh}
Andoni, A. and Indyk, P.
\newblock E2lsh: Exact euclidean locality-sensitive hashing.
\newblock \emph{Technical report}, 2004.

\bibitem[Bamler \& Mandt(2020)Bamler and Mandt]{bamler2020extreme}
Bamler, R. and Mandt, S.
\newblock Extreme classification via adversarial softmax approximation.
\newblock In \emph{International Conference on Learning Representations}, 2020.
\newblock URL \url{https://openreview.net/forum?id=rJxe3xSYDS}.

\bibitem[Bengio et~al.(2019)Bengio, Dembczynski, Joachims, Kloft, and
  Varma]{bengio2019extreme}
Bengio, S., Dembczynski, K., Joachims, T., Kloft, M., and Varma, M.
\newblock Extreme classification (dagstuhl seminar 18291).
\newblock Schloss Dagstuhl-Leibniz-Zentrum fuer Informatik, 2019.

\bibitem[Bengio \& Sen{\'e}cal(2008)Bengio and Sen{\'e}cal]{bengio2008adaptive}
Bengio, Y. and Sen{\'e}cal, J.-S.
\newblock Adaptive importance sampling to accelerate training of a neural
  probabilistic language model.
\newblock \emph{IEEE Transactions on Neural Networks}, 19\penalty0
  (4):\penalty0 713--722, 2008.

\bibitem[Bhatia et~al.(2016)Bhatia, Dahiya, Jain, Mittal, Prabhu, and
  Varma]{Bhatia16}
Bhatia, K., Dahiya, K., Jain, H., Mittal, A., Prabhu, Y., and Varma, M.
\newblock The extreme classification repository: Multi-label datasets and code,
  2016.
\newblock URL \url{http://manikvarma.org/downloads/XC/XMLRepository.html}.

\bibitem[Charikar \& Siminelakis(2017)Charikar and
  Siminelakis]{charikar2017hashing}
Charikar, M. and Siminelakis, P.
\newblock Hashing-based-estimators for kernel density in high dimensions.
\newblock In \emph{2017 IEEE 58th Annual Symposium on Foundations of Computer
  Science (FOCS)}, pp.\  1032--1043. IEEE, 2017.

\bibitem[Chen \& Shrivastava(2018)Chen and Shrivastava]{chen2018densified}
Chen, B. and Shrivastava, A.
\newblock Densified winner take all (wta) hashing for sparse datasets.
\newblock In \emph{Uncertainty in artificial intelligence}, 2018.

\bibitem[Chen et~al.(2018)Chen, Shrivastava, Steorts, et~al.]{chen2018unique}
Chen, B., Shrivastava, A., Steorts, R.~C., et~al.
\newblock Unique entity estimation with application to the syrian conflict.
\newblock \emph{The Annals of Applied Statistics}, 12\penalty0 (2):\penalty0
  1039--1067, 2018.

\bibitem[Chen et~al.(2019{\natexlab{a}})Chen, Medini, Farwell, Gobriel, Tai,
  and Shrivastava]{chen2019slide}
Chen, B., Medini, T., Farwell, J., Gobriel, S., Tai, C., and Shrivastava, A.
\newblock Slide : In defense of smart algorithms over hardware acceleration for
  large-scale deep learning systems, 2019{\natexlab{a}}.

\bibitem[Chen et~al.(2019{\natexlab{b}})Chen, Xu, and Shrivastava]{beidi_sgd}
Chen, B., Xu, Y., and Shrivastava, A.
\newblock Fast and accurate stochastic gradient estimation.
\newblock In Wallach, H., Larochelle, H., Beygelzimer, A., d\textquotesingle
  Alch\'{e}-Buc, F., Fox, E., and Garnett, R. (eds.), \emph{Advances in Neural
  Information Processing Systems 32}, pp.\  12339--12349. Curran Associates,
  Inc., 2019{\natexlab{b}}.

\bibitem[Chen et~al.(2019{\natexlab{c}})Chen, Xu, and Shrivastava]{chen2018lsh}
Chen, B., Xu, Y., and Shrivastava, A.
\newblock Fast and accurate stochastic gradient estimation.
\newblock In \emph{Advances in Neural Information Processing Systems}, pp.\
  12339--12349, 2019{\natexlab{c}}.

\bibitem[Choromanska \& Langford(2015)Choromanska and
  Langford]{NIPS2015_e369853d}
Choromanska, A.~E. and Langford, J.
\newblock Logarithmic time online multiclass prediction.
\newblock In Cortes, C., Lawrence, N., Lee, D., Sugiyama, M., and Garnett, R.
  (eds.), \emph{Advances in Neural Information Processing Systems}, volume~28,
  pp.\  55--63. Curran Associates, Inc., 2015.
\newblock URL
  \url{https://proceedings.neurips.cc/paper/2015/file/e369853df766fa44e1ed0ff613f563bd-Paper.pdf}.

\bibitem[Coleman \& Shrivastava(2020)Coleman and Shrivastava]{coleman2020sub}
Coleman, B. and Shrivastava, A.
\newblock Sub-linear race sketches for approximate kernel density estimation on
  streaming data.
\newblock In \emph{Proceedings of The Web Conference 2020}, pp.\  1739--1749,
  2020.

\bibitem[Coleman et~al.(2019)Coleman, Baraniuk, and
  Shrivastava]{coleman2019sub}
Coleman, B., Baraniuk, R.~G., and Shrivastava, A.
\newblock Sub-linear memory sketches for near neighbor search on streaming
  data.
\newblock \emph{arXiv preprint arXiv:1902.06687}, 2019.

\bibitem[Daghaghi et~al.(2021)Daghaghi, Meisburger, Zhao, and
  Shrivastava]{daghaghi2021accelerating}
Daghaghi, S., Meisburger, N., Zhao, M., and Shrivastava, A.
\newblock Accelerating slide deep learning on modern cpus: Vectorization,
  quantizations, memory optimizations, and more.
\newblock \emph{Proceedings of Machine Learning and Systems}, 3, 2021.

\bibitem[Dong et~al.(2018)Dong, Xu, and Xu]{dong2018speech}
Dong, L., Xu, S., and Xu, B.
\newblock Speech-transformer: a no-recurrence sequence-to-sequence model for
  speech recognition.
\newblock In \emph{2018 IEEE International Conference on Acoustics, Speech and
  Signal Processing (ICASSP)}, pp.\  5884--5888. IEEE, 2018.

\bibitem[Gutmann \& Hyvärinen(2010)Gutmann and Hyvärinen]{pmlr-v9-gutmann10a}
Gutmann, M. and Hyvärinen, A.
\newblock Noise-contrastive estimation: A new estimation principle for
  unnormalized statistical models.
\newblock In Teh, Y.~W. and Titterington, M. (eds.), \emph{Proceedings of the
  Thirteenth International Conference on Artificial Intelligence and
  Statistics}, volume~9 of \emph{Proceedings of Machine Learning Research},
  pp.\  297--304, Chia Laguna Resort, Sardinia, Italy, 13--15 May 2010. PMLR.
\newblock URL \url{http://proceedings.mlr.press/v9/gutmann10a.html}.

\bibitem[Indyk \& Motwani(1998)Indyk and Motwani]{indyk1998approximate}
Indyk, P. and Motwani, R.
\newblock Approximate nearest neighbors: towards removing the curse of
  dimensionality.
\newblock In \emph{Proceedings of the thirtieth annual ACM symposium on Theory
  of computing}, pp.\  604--613, 1998.

\bibitem[Indyk \& Woodruff(2006)Indyk and Woodruff]{indyk2006polylogarithmic}
Indyk, P. and Woodruff, D.
\newblock Polylogarithmic private approximations and efficient matching.
\newblock In \emph{Theory of Cryptography Conference}, pp.\  245--264.
  Springer, 2006.

\bibitem[Jain et~al.(2019)Jain, Balasubramanian, Chunduri, and
  Varma]{jain2019slice}
Jain, H., Balasubramanian, V., Chunduri, B., and Varma, M.
\newblock Slice: Scalable linear extreme classifiers trained on 100 million
  labels for related searches.
\newblock In \emph{Proceedings of the Twelfth ACM International Conference on
  Web Search and Data Mining}, pp.\  528--536, 2019.

\bibitem[Jean et~al.(2014)Jean, Cho, Memisevic, and Bengio]{jean2014using}
Jean, S., Cho, K., Memisevic, R., and Bengio, Y.
\newblock On using very large target vocabulary for neural machine translation.
\newblock \emph{arXiv preprint arXiv:1412.2007}, 2014.

\bibitem[Langley(2000)]{langley00}
Langley, P.
\newblock Crafting papers on machine learning.
\newblock In Langley, P. (ed.), \emph{Proceedings of the 17th International
  Conference on Machine Learning (ICML 2000)}, pp.\  1207--1216, Stanford, CA,
  2000. Morgan Kaufmann.

\bibitem[Luo \& Shrivastava(2019)Luo and Shrivastava]{luo2019scaling}
Luo, C. and Shrivastava, A.
\newblock Scaling-up split-merge mcmc with locality sensitive sampling (lss).
\newblock In \emph{Proceedings of the AAAI Conference on Artificial
  Intelligence}, volume~33, pp.\  4464--4471, 2019.

\bibitem[Medini et~al.(2019)Medini, Huang, Wang, Mohan, and
  Shrivastava]{medini2019extreme}
Medini, T. K.~R., Huang, Q., Wang, Y., Mohan, V., and Shrivastava, A.
\newblock Extreme classification in log memory using count-min sketch: A case
  study of amazon search with 50m products.
\newblock In \emph{Advances in Neural Information Processing Systems}, pp.\
  13244--13254, 2019.

\bibitem[Mikolov et~al.(2013)Mikolov, Sutskever, Chen, Corrado, and
  Dean]{mikolov2013distributed}
Mikolov, T., Sutskever, I., Chen, K., Corrado, G.~S., and Dean, J.
\newblock Distributed representations of words and phrases and their
  compositionality.
\newblock In \emph{Advances in neural information processing systems}, pp.\
  3111--3119, 2013.

\bibitem[Owens et~al.(2008)Owens, Houston, Luebke, Green, Stone, and
  Phillips]{owens2008gpu}
Owens, J.~D., Houston, M., Luebke, D., Green, S., Stone, J.~E., and Phillips,
  J.~C.
\newblock Gpu computing.
\newblock 2008.

\bibitem[Pennington et~al.(2014)Pennington, Socher, and
  Manning]{pennington2014glove}
Pennington, J., Socher, R., and Manning, C.~D.
\newblock Glove: Global vectors for word representation.
\newblock In \emph{Proceedings of the 2014 conference on empirical methods in
  natural language processing (EMNLP)}, pp.\  1532--1543, 2014.

\bibitem[Rawat et~al.(2019)Rawat, Chen, Yu, Suresh, and
  Kumar]{rawat2019sampled}
Rawat, A.~S., Chen, J., Yu, F. X.~X., Suresh, A.~T., and Kumar, S.
\newblock Sampled softmax with random fourier features.
\newblock In \emph{Advances in Neural Information Processing Systems}, pp.\
  13834--13844, 2019.

\bibitem[Shrivastava \& Li(2014)Shrivastava and Li]{shrivastava2014asymmetric}
Shrivastava, A. and Li, P.
\newblock Asymmetric lsh (alsh) for sublinear time maximum inner product search
  (mips).
\newblock In \emph{Advances in Neural Information Processing Systems}, pp.\
  2321--2329, 2014.

\bibitem[Spring \& Shrivastava(2017{\natexlab{a}})Spring and
  Shrivastava]{spring2017new}
Spring, R. and Shrivastava, A.
\newblock A new unbiased and efficient class of lsh-based samplers and
  estimators for partition function computation in log-linear models.
\newblock \emph{arXiv preprint arXiv:1703.05160}, 2017{\natexlab{a}}.

\bibitem[Spring \& Shrivastava(2017{\natexlab{b}})Spring and
  Shrivastava]{spring2017scalable}
Spring, R. and Shrivastava, A.
\newblock Scalable and sustainable deep learning via randomized hashing.
\newblock In \emph{Proceedings of the 23rd ACM SIGKDD International Conference
  on Knowledge Discovery and Data Mining}, pp.\  445--454, 2017{\natexlab{b}}.

\bibitem[Spring \& Shrivastava(2020)Spring and Shrivastava]{spring2020mutual}
Spring, R. and Shrivastava, A.
\newblock Mutual information estimation using lsh sampling.
\newblock In \emph{Proceedings of the 29th International Joint Conference on
  Artificial Intelligence, AAAI Press}, 2020.

\bibitem[Vijayanarasimhan et~al.(2014)Vijayanarasimhan, Shlens, Monga, and
  Yagnik]{vijayanarasimhan2014deep}
Vijayanarasimhan, S., Shlens, J., Monga, R., and Yagnik, J.
\newblock Deep networks with large output spaces.
\newblock \emph{arXiv preprint arXiv:1412.7479}, 2014.

\bibitem[Wang et~al.(2017)Wang, Jiang, Qian, Yang, Li, Zhang, Wang, and
  Tang]{wang2017residual}
Wang, F., Jiang, M., Qian, C., Yang, S., Li, C., Zhang, H., Wang, X., and Tang,
  X.
\newblock Residual attention network for image classification.
\newblock In \emph{Proceedings of the IEEE conference on computer vision and
  pattern recognition}, pp.\  3156--3164, 2017.

\bibitem[Yagnik et~al.(2011)Yagnik, Strelow, Ross, and Lin]{yagnik2011power}
Yagnik, J., Strelow, D., Ross, D.~A., and Lin, R.-s.
\newblock The power of comparative reasoning.
\newblock In \emph{2011 International Conference on Computer Vision}, pp.\
  2431--2438. IEEE, 2011.

\bibitem[Yao et~al.(2019)Yao, Mao, and Luo]{yao2019graph}
Yao, L., Mao, C., and Luo, Y.
\newblock Graph convolutional networks for text classification.
\newblock In \emph{Proceedings of the AAAI Conference on Artificial
  Intelligence}, volume~33, pp.\  7370--7377, 2019.

\bibitem[Zhang et~al.(2015)Zhang, Zhao, and LeCun]{zhang2015character}
Zhang, X., Zhao, J., and LeCun, Y.
\newblock Character-level convolutional networks for text classification.
\newblock In \emph{Advances in neural information processing systems}, pp.\
  649--657, 2015.

\end{thebibliography}
\bibliographystyle{icml2021}

\newpage
\appendix
\section{DWTA Hash: Densified Winner Take All Hash} \label{appendix: Dwta}
DWTA \citep{chen2018densified}  hash maps the data into a transformed space such that the hamming distance between vectors correlates with their rank similarity measure in the original space. Traditional WTA hashing \citep{yagnik2011power} works by taking a vector $x$ and applying a permutation $\Theta$. The hash value is then taken as the index of the largest of the first $R$ values in the permutation $\Theta(x)$. From here we can use the standard $(K, L)$ LSH parameterization by simply computing the hash on $K \times L$ permutations of the vector, or more efficiently, group a single permutation into $K\times L$ bins and computing the hash as the index of the largest component in each bin. The intuition behind this hashing scheme is that it tends to group vectors based on their largest components, which is useful in this type of learning application in which we are interested in maximizing the inner product between two vectors. Densified WTA hashing (DWTA) is an extension to standard WTA hashing to improve its discriminative power on sparse datasets. Classical WTA may fail on sparse datasets because it becomes likely that some bin of the permuted vector contains no non-zero terms. DWTA uses densification to solve this problem by looking at the bins that contain no non-zeros and taking their hash to be the hash of the nearest bin that does contain a non-zero component. It has been shown \citep{chen2018densified}  that the collision probability of DWTA is precisely the collision probability of WTA hash for nonempty bins, irrespective of the sparsity.

\section{Dataset}
\label{appendix:dataset}
\textbf{Amazon-670K} dataset is a product recommendation dataset with 670K labels. Each input is a vector representation of a product, and the corresponding labels are other products (among 670K choices) that a user might be interested in purchase. This is an anonymized and aggregated behavior data from Amazon and poses a signiﬁcant challenge owing to a large number of classes.\\
\textbf{Wiki-325K} dataset is extracted from Wikipedia and consists of over 1.7 Million training samples, 1.6 Million sparse feature dimension and 325K labels correspond to the category of the instance.\\
\textbf{ODP} is extracted from Open Directory Project, a comprehensive human-edited directory of the Web. Each sample in the dataset is a document,
and the feature representation is bag-of-words. The class label is the category associated with the document.\\
\textbf{Amazon-Uniform} is a subsampled version of Amazon-670K and its label distribution is near uniform. More details are in Section \ref{sec: powerlaw}.
\section{Sparsity}
The number of negative samples $C$ is more like a  budget hyperparameter for sparsity e.g. for Amazon-670K, we set sparsity 0.05, meaning $C = 0.05N$, and we keep sampling the buckets from hash tables till the sparsity budget is exhausted and then we stop.
Generally, C is independent of $N$. For example, Wiki-325K has 325k classes and it requires more sparsity (0.1) compared to Amazon-670K which has twice the $N$ (670k) where 0.05 sparsity is sufficient to get the best accuracy. $C$ is dependent on the dataset and the hardness of classification. 
Also for ODP-105K dataset $C=0.04N$, and for Amazon-Uniform dataset $C=0.02N$. Table \ref{table:Cimpact} shows that both versions of our algorithm performs surprisingly well even with extremely low sparsity \textbf{(0.005)}, while static-based methods such as Sampled softmax enormously fails, and it requires \textbf{at least 0.2} sparsity to hit \textbf{even 35\%} accuracy. 
\begin{table}[H]
    \centering
    \small
    \caption{Impact of number of negative samples $C$ on P@1(\%) for Amazon-670K dataset.}
    \vskip 0.05in
    \begin{tabular}{c|c|c|c} \hline
    	 $C$  & LSH Embedding & LSH Label& Sampled Softmax  \\ \hline
    	$0.005N$ & 33.4  & 32 &16 \\ \hline
    $0.05N$ & 36.1 &  35.5& 32.4\\ \hline
    	\hline
    	\end{tabular}
    	\label{table:Cimpact}
\end{table}

\subsection{Impact of Number of Hash Tables}
Tables \ref{table:Hash_Num_amz670} and \ref{table:Hash_Num_Wiki} show the effect of the number of hash tables $L$ on P@1 for Amazon-670K and Wiki-325K datasets.  Increasing number of hash tables helps the algorithm to retrieve the informative samples with higher probability, thus improves the accuracy. However, it increases the average training time per epoch proportionally.  

\begin{table}[H]
    \centering
    \footnotesize
    \caption{Impact of Number of Hash Tables on
    Amazon-670K}
    \begin{tabular}{c|c|c} \hline
    	 Num of Hash Tables  & P@1(\%) & 
    	 Avg training time per epoch\\ \hline
    	$L=100$ & 34.5  & Baseline \\ \hline
     $L=200$ & 35.4 &  0.54x\\ \hline
    $L=300$ & 35.7 &  0.36x\\ \hline
    	$L=400$ & 36 & 0.26x\\
    	\hline
    		\bottomrule
    	\end{tabular}
    	\label{table:Hash_Num_amz670}
\end{table}

\begin{table}[H]
    \centering
    \small
    \caption{Impact of Number of Hash Tables on Wiki-325K}
    \begin{tabular}{c|c|c} \hline
    	 Num of Hash Tables  & P@1(\%) & Avg training time per epoch  \\ \hline
        $L=100$ & 53.4  & Baseline\\ \hline
         $L=200$ & 55.4 &  0.60x\\ \hline
    $L=300$ & 56.3 & 0.47x\\
    
    	\hline
    		\bottomrule
    	\end{tabular}
    	\label{table:Hash_Num_Wiki}
    
\end{table}

There is a trade-off between increasing L and running time. 
Increasing L improves the accuracy, however at some point it reaches a saturation mode where the accuracy improvement is slight comparing to the increase in time. 

   


\subsection{Impact of Number of Hash Functions}
Table \ref{table:Hash_Num} shows the impact of the number of hash functions on accuracy for all datasets with the LSH Embedding scheme. The number of hash functions $K$ determines sampling quality. LNS with an optimal value of $K$ retrieves informative samples, while smaller $K$ retrieves less informative and low-quality samples, and larger values of $K$ leads the algorithm to miss the important and informative samples. For instance, the optimal number of hash functions for Wiki-325K is $K=5$ which achieves higher accuracy than $K=4$ or $K=6$.

\begin{table}[H]
    \centering
    \footnotesize
    
    \caption{ Impact of number of hash functions $K$ on P@1(\%)}
    \vskip 0.10in
    \begin{tabular}{c|c|c|c}
     \toprule
    \hline
    Dataset	&$K=4$  & $K=5$ & $K=6$  \\ \hline
    	Amazon-670 & -  & 33.4 & 36.1 \\ \hline
   Wiki-325K & 38.4 &  56.3 & 54.6\\ \hline
    ODP-105K & 16.5 &  16.8 & 15.4\\ \hline
    	Amaz-Uniform & 22.1 & 23.1 & 22.8\\
    	\hline
    \bottomrule
    	\end{tabular}
    	\label{table:Hash_Num}
\end{table}


\end{document}